\documentclass{article}

\usepackage[english]{babel}
\usepackage[utf8]{inputenc}
\usepackage[letterpaper,top=2cm,bottom=2cm,left=3cm,right=3cm,marginparwidth=1.75cm]{geometry}
\usepackage{authblk} 
\usepackage{amsmath, amssymb}
\usepackage{graphicx}
\usepackage{listings}
\usepackage{xcolor}
\usepackage[colorlinks=true, allcolors=blue]{hyperref}

\usepackage[style=numeric, sorting=none]{biblatex}
\definecolor{codegreen}{rgb}{0,0.6,0}
\definecolor{codegray}{rgb}{0.5,0.5,0.5}
\definecolor{codepurple}{rgb}{0.58,0,0.82}
\definecolor{backcolour}{rgb}{0.95,0.95,0.92}
\definecolor{codeblue}{rgb}{0.0,0.0,0.8}
\lstdefinestyle{pythonstyle}{
    backgroundcolor=\color{backcolour},   
    commentstyle=\color{codegreen},
    keywordstyle=\color{codeblue},
    numberstyle=\tiny\color{codegray},
    stringstyle=\color{codepurple},
    basicstyle=\ttfamily\small,
    breakatwhitespace=false,         
    breaklines=true,                 
    captionpos=b,                    
    keepspaces=true,                 
    numbers=left,                    
    numbersep=5pt,                  
    showspaces=false,                
    showstringspaces=false,
    showtabs=false,                  
    tabsize=4,
    upquote=true,
    basicstyle=\ttfamily,
    language=Python
}
\title{PyCC.id: A package for hypothesis-driven equation discovery with structural identifiability} 
\author{Federico J. Gonzalez}
\affil{Physics Institute of Rosario (IFIR), CONICET-UNR, Blvd. 27 de Febrero 210 Bis, Rosario, S2000EZP, Argentina.}
\affil{E-mail: fgonzalez@ifir-conicet.gov.ar. ORCID: 0000-0003-2026-4129}

\begin{document}
\maketitle


The software package is available at: \url{https://github.com/FedejGon/pyCC.id}

\section*{Summary}

Data-driven equation discovery is fundamentally an inverse problem that seeks to infer the governing differential equations of a system directly from time-series measurements. A known issue is the ill-conditioned nature of the inverse problem, which frequently produces multiple mathematical models that fit the data similarly well. One path to address this issue is by incorporating known hypotheses and constraints into the training phase beforehand. While this approach effectively reduces the search space, it still results in multiple candidate models, forcing practitioners to rely on post-hoc manual filtering based on their own domain expertise. 
A recent approach incorporates structural `skeletons' inspired by characteristic curves (CCs), defining a hypothesis-driven methodology. In this methodology, practitioners define a skeleton, which is associated with a family of ordinary differential equations (ODEs), and then add their hypotheses and priors based on their domain knowledge to refine the obtained model iteratively. An important advantage of this approach is that some skeletons have demonstrable structural identifiability properties, which are useful for checking whether the skeleton is correct or should be discarded. Furthermore, this formalism enables the use of multiple equation discovery paradigms due to its modularity (such as neural networks, symbolic regression, and sparse regression).  
In this work, we present the Python library \texttt{PyCC}, which condenses these efforts into a flexible tool that allows researchers and engineers to seamlessly define their skeletons and hypotheses to discover ODEs from time-dependent data.

\section*{Statement of need}

Data-driven equation discovery is fundamentally an inverse problem that seeks to infer the 
governing differential equations of a system from time-series measurements. 
Historically, this topic is rooted in the field of system identification within control theory. 
However, modern equation discovery methods have introduced novel developments that have enabled applications across a wide range of disciplines in science, medicine, and engineering. 
During the last two decades, this interdisciplinary effort has contributed significantly to the emerging field of scientific machine learning\cite{Dietrich2025}.

Despite these advances, the inverse problem remains typically ill-conditioned\cite{Schonlieb2025}, as many different mathematical expressions can fit a finite dataset with similar accuracy. Consequently, the algorithms often provide a vast landscape of candidate models. 
Although these models are valid in terms of data fitting, they often vary significantly in their structural form, leading to physically inconsistent representations. 
Therefore, practitioners must often rely on post-hoc analysis to manually discard models that do not align with the physics based on their domain expertise. This leaves the final selection of a model somewhat arbitrary.


To address this difficulty, an emerging method incorporates the concepts of structural `skeletons' and CCs\cite{Gonzalez2023,Gonzalez2024,Gonzalez2025,Gonzalez2026}. An structural skeleton corresponds to a given classification of families of ODEs. The choice of the structural skeleton directly affects identifiability: while certain skeletons possess structural identifiability, others can result in non-identifiable or ambiguous representations. When a skeleton is shown to be theoretically identifiable in phase space \cite{Gonzalez2026}, it provides a formal framework to determine whether a proposed equation is consistent with the data or should be reformulated.

This work introduces \texttt{PyCC.id} (also referred to as \texttt{PyCC}), an open-source library that implements this framework through a hypothesis-driven methodology. This hypothesis-driven workflow requires the practitioner to propose a specific structural skeleton based on physical intuition or domain knowledge. \texttt{PyCC} then allows the user to integrate prior information (such as explicit physical symmetries and constraints) directly into the functions defined within these skeletons. By providing a clear pathway to validate or eliminate these hypothesized structures at the algorithmic level, the library contributes to addressing the ill-conditioned nature of the inverse problem. 
It enables practitioners to easily define skeletons and additional hypotheses that sufficiently reduce the search space to mitigate structural ambiguity, facilitating the discovery of interpretable and physically consistent governing equations.

\section*{Structural skeletons}
The \texttt{PyCC} library operates by enforcing a structural skeleton, which serves as a template to define a family of admissible physical models. Unlike purely data-driven `black-box' methods, this hypothesis-driven approach constrains the model to a physically motivated structure, ensuring that the identified components are structurally identifiable and grounded in the underlying physics.

The library decomposes global dynamics into these skeletons composed of univariate one-dimensional (1D) functions. The following examples are skeletons with structural identifiability properties\cite{Gonzalez2026}:

\begin{enumerate}
\item \textbf{A first-order family:} With applications in overdamped systems such as mechanical nonlinear damping, viscoelastic materials, nonlinear RL series and RC parallel circuits\cite{Gonzalez2023,Gonzalez2024}.
    \begin{equation}
        F_{ext}(t) = f_1(x) + f_2(x)\dot{x}
    \end{equation}

\item \textbf{A second-order family with velocity-dependent friction:} The standard model for mechanical oscillators with velocity-dependent friction, also known as a generalized Rayleigh-type nonlinear oscillation with velocity-dependent friction and external forcing)\cite{Gonzalez2025,Gonzalez2026}.
    \begin{equation}
        F_{ext}(t) = \ddot{x} + f_1(\dot{x}) + f_2(x)
        \label{ec:velocity-dependent}
    \end{equation}

\item \textbf{A second-order family with position-dependent friction:} Designed for systems where damping varies with position, also known as a Liénard equation with external forcing\cite{Gonzalez2025}.
    \begin{equation}
        F_{ext}(t) = \ddot{x} + f_1(x)\dot{x} + f_2(x)
    \end{equation}

\item \textbf{A coupled multi-degree-of-freedom (MDOF) family:} Modeling the interactions for a coupled system\cite{Gonzalez2026,Nayfeh_2004}.
    \begin{align}
        F_{ext,1}(t) &= \ddot{x}_1 + f_1(\dot{x}_1) + f_2(x_1 - x_2) \\
        F_{ext,2}(t) &= \ddot{x}_2 + f_3(\dot{x}_2) + f_4(x_2 - x_1)        
    \end{align}

\end{enumerate}

The library enables the user to seamlessly define their own proposed skeletons while also integrating specific physical priors directly into the $f_i$ functions, such as enforcing odd/even parity for $f_1$, or specifying values like $f_2(1)=0$.
The $f_i$ functions can be strategically defined to match the constitutive relations or CCs of the system. For instance, in the velocity-dependent friction model (Eq.~\ref{ec:velocity-dependent}), $f_1$ and $f_2$ are associated with dissipative (friction) and elastic (stiffness) elements, respectively.  

Depending on the domain knowledge of the practitioner and the desired level of interpretability, these 1D functions can be identified using three parametrizations:

\begin{itemize}
\item \textbf{Functional ($\{f_i\}; \;i=0\,,1\,,\ldots$)}: The 1D functions are treated as unknown functions and approximated, for instance, using universal approximators such as Neural Networks (NN) or Symbolic Regression (SymbR).
\item \textbf{Parametric ($\{a_i\}; \;i=0\,,1\,,\ldots$)}: The functional form is completely hypothesized and parameterized using scalar parameters $\{a_i\}$. The library identifies the optimal values for $\{a_i\}$ through nonlinear optimization. 
\item \textbf{Hybrid ($\{f_i\},\{a_j\}; \;i,j=0\,,1\,,\ldots$)}: This approach enables practitioners to simultaneously fit 1D functions $\{f_i\}$ and parameters $\{a_j\}$.
\end{itemize}

\section*{Relation to other packages}

\texttt{PyCC} can be considered as an integrative framework within the scientific machine learning ecosystem.
Its primary differentiator is a hypothesis-driven workflow that shifts the focus from purely data-driven discovery to a structural `skeleton' approach, ensuring physical consistency and structural identifiability. While many packages seek to find the `best' equation from scratch, this library provides a formal pipeline to validate if a hypothesized physical structure is consistent with the measured dynamics. 

The library is not intended to replace or compete with established packages, but rather to serve as an interface that uses their strengths (such as the symbolic power of \texttt{PySR} and the flexibility of neural networks) within a structured framework that prioritizes identifiability, interpretability, and physical consistency.

\begin{itemize}
\item Sparse identification (e.g., \texttt{SINDy}\cite{brunton2016}): This approach is highly effective when the system dynamics can be represented as a sparse combination of terms from a pre-defined library. However, its success is highly dependent on the user correctly guessing the necessary basis functions. If a complex or non-standard term (such as a specific friction model or a saturation curve) is absent, the results may be misleading. While the CC-based formalism can theoretically be implemented via sparse regression, the \texttt{PyCC} library focuses on reconstructing the \textit{shape} of unknown constitutive relations directly. By using, for instance, universal approximators like NNs or SymbR, \texttt{PyCC} captures arbitrary functional forms without requiring a rigid prior library of basis terms. 


\item Symbolic regression (e.g., \texttt{PySR}\cite{Cranmer2023PySR} ): 
Standard symbolic regression tools often require searching vast spaces to discover relationships in multivariate data. 
In contrast, \texttt{PyCC} acts as a manager that internally uses \texttt{PySR} through two specialized workflows: 
\begin{itemize}
\item \textit{Iterative symbolic discovery:} This method decomposes the multivariate problem into a sequence of simpler, 1D symbolic regression that iteratively minimize the residual according to the user-defined skeleton. 
\item \textit{Symbolic post-processing:} The library features an automated post-processing tool to extract analytical expressions from identified CCs (obtained, for instance, via the NN method). This hybrid strategy has demonstrated superior robustness to noise compared to purely NN-based identification\cite{Gonzalez2026}.
\end{itemize}

\item Neural ODEs: Standard implementations \cite{Chen2018} usually learn entire vector fields using a single, monolithic NN. While powerful for prediction, these `black-box' models offer limited physical insight. \texttt{PyCC} adopts a `grey-box' philosophy by modeling only the specific, unknown constitutive relations within a fixed structural skeleton. This ensures that the learned components (such as damping or restoring elements) retain a clear, isolated physical interpretation regardless of the numerical backend used. 

\item Physics-informed neural networks (PINNs): \texttt{PINNs} \cite{Raissi2019} excel at solving forward and inverse problems by embedding physical laws into a loss function. However, they are often less suited for discovery when the functional forms of interactions are entirely unknown. \texttt{PyCC} differs by explicitly isolating these unknown terms as distinct functions ($f_i$) to be learned, making them directly accessible for independent visualization, analysis, and physical validation.

    
\end{itemize}

\section*{Features}

\texttt{PyCC} is designed to be a user-friendly and highly-customizable tool for researchers and practitioners. Its key features include:

\begin{itemize}
\item Interpretable models: It allows practitioners to add prior information by decomposing the complex, high-dimensional functional space into a set of simple, one-dimensional CCs that have direct physical meaning (i.e., the CCs are directly the constitutive relations of the system elements, such as, stiffness or damping), and a set of scalar parameters (e.g., mass).

\item  Flexible function parameterization: Supports multiple back-ends for modeling the 1D functions, allowing practitioners to switch between different representations with minimal modifications:
\begin{itemize}
    \item Neural Networks (NN): Implemented using \texttt{PyTorch} \cite{Paszke2019} and compatible with  \texttt{GPUs} and multi-core \texttt{CPUs}. Natively supports both NVIDIA (`cuda') and Intel (`xpu' via `intel-extension-for-pytorch') \texttt{GPUs} for training the NNs.
    \item Polynomials (Poly): Provides an expansion of the $f_i$ functions using polynomial basis functions.
    \item Symbolic regression (SymbR): Utilizes the \texttt{PySR} package \cite{Cranmer2023PySR} to discover analytical expressions that are compatible with a given prior structure, while also serving as a post-processing tool.   
\end{itemize}
\item Physics-informed discovery: Allows users to inject domain knowledge as constraints during training (e.g., `f1 odd', and `f2(0)=0') or by defining conserved quantities that are added to the loss function. This leads to more robust and physically consistent models. 
\item Hardware acceleration: Natively supports multicore CPUs and GPUs from both NVIDIA (`cuda') and Intel (`xpu' via `intel-extension-for-pytorch') for training the NNs.

\item Built-in simulator: Includes a versatile ODE solver (`pycc.simulate') compatible with all identification methods. This solver facilitates forward integrations of both the identified models and the theoretical equations used to generate training databases.
\item Comprehensive documentation: Provides a quick-start Google Colab tutorial with an accompanying YouTube video, along with a complete documentation, examples, and recommended workflows. 
\end{itemize}

\section*{Core functions and methods}
This section briefly describes the three main methods that are defined in the \texttt{PyCC} library: \texttt{pycc.simulate()}, \texttt{pycc.train()}, and \texttt{pycc.post\_processing()}.

\subsection*{\texttt{1. pycc.simulate()}} \vspace{-0.1 cm}
This function is responsible for forward system integrations over time. It acts as a simulation manager, dispatching the task to different methods (such as `Theoretical', `NN', `SymbR', `Poly', or `Interp'). It has two main uses:
\vspace{-0.2cm}
\begin{itemize}
    \item \textit{Theoretical simulation:} Can be used to integrate forward the theoretical equations in order to obtain ground-truth datasets that will be used for training the models, or to generate theoretical forward integration under different initial conditions or driven forces to compare against model predictions. 
 \vspace{-0.2cm}   \item \textit{Validation:} Integrates discovered models after training (e.g., from a NN model) to verify their accuracy against expected dynamics.
\end{itemize}

\subsection*{\texttt{2. pycc.train()}} \vspace{-0.1 cm}
This function serves as the primary entry point for identifying system dynamics from a given dataset. It operates as a training dispatcher that abstracts the complexity of individual algorithms, automatically routing data to the appropriate module based on the specified `method' parameter (such as `NN', `SymbR', `Poly'). 

\subsection*{\texttt{3. pycc.post\_processing()}} \vspace{-0.1 cm}
This acts as a standalone utility, specifically designed to obtain analytical expressions for the obtained CCs. It converts numerical CCs (previously obtained from methods such as `NN' or `Poly') into explicit symbolic expressions. It has two main uses: 
\vspace{-0.2cm}
\begin{itemize}
\item \textit{Model conversion:} Transforms the obtained CCs into symbolic functions for inspection, analysis, and interpretation. 
\vspace{-0.2cm}    \item \textit{Pre-simulation preparation:} Generates optimized symbolic models that are fully compatible with \texttt{pycc.simulate()} for subsequent simulations.
\end{itemize}

\section*{Illustrative example: A second-order ODE}
To demonstrate the capabilities of \texttt{PyCC}, we consider a classic physical system: a nonlinear oscillator with friction. The governing equation is a second-order differential equation:

\begin{equation}
\ddot{x} + \delta\dot{x} + \mu\tanh(500\dot{x}) + \alpha x + \beta x^3 = F_{ext}(t)
\end{equation}

where $F_{ext}(t) = A\cos(\omega t)$ is an external driving force. The term $\tanh(500\dot{x})$ acts as a smooth approximation of the signum function, $\text{sign}(\dot{x})$, effectively modeling Coulomb friction.

To apply the \texttt{PyCC} framework, we rewrite this system as a set of first-order ordinary differential equations (ODEs). By defining the state variables $x_1 = x$ (position) and $x_2 = \dot{x}$ (velocity), the system becomes:

\begin{equation}
\begin{cases}
\dot{x}_1 = x_2 \\
\dot{x}_2 = F_{ext}(t) - \delta x_2 - \mu\tanh(500x_2) - \alpha x_1 - \beta x_1^3
\end{cases}
\end{equation}

The system is simulated with defined parameters and initial conditions to generate the dataset $\mathcal{D} = \{x_1, x_2, \dot{x}_1, \dot{x}_2, F_{ext}\}$, which provides the input for system identification. 
The following code shows how to use the \texttt{PyCC} library to integrate these theoretical equations in order to generate the input dataset that will be used later for training the models:

\begin{lstlisting}
import pycc
import numpy as np
import pandas as pd
import matplotlib.pyplot as plt

##############################################
# Simulating a stick-slip second order system using 
# pycc.simulate(method='Theoretical')
# Defining parameters and theoretical functions
alpha=1.0;beta=0.2;delta=0.1;Omega=1.0;
x0=0.0;v0=0.0; y0=[x0,v0] # initial conditions
t_span=(0, 20); t_eval=np.linspace(*t_span, 1000)
def F1_th(x_dot):
    return delta * x_dot + 0.5 * np.tanh(500*x_dot)
def F2_th(x):
    return alpha * x + beta * x**3
def F_ext(t):
    return np.cos(Omega * t)
# Defining the theoretical equation
eqs_th = ['x1_dot = x2',
          'x2_dot = F_ext - f1(x2) - f2(x1)']
# Defining the simulation parameters
params_th = {
    't_span': t_span,
    'y0': y0,
    't_eval': t_eval,
    'method': 'LSODA',
    'local_funcs': {'f1': lambda t: F1_th(t),
                    'f2': lambda t: F2_th(t),'F_ext': lambda t: F_ext(t)}
}
# Integrating forward the theoretical equation
sol,derivatives = pycc.simulate(eqs_th,method='Theoretical', params=params_th)
# Extracting data 
time_data    = sol.t
x1_data      = sol.y[0]
x2_data      = sol.y[1]
x1_dot_data  = derivatives[0]
x2_dot_data  = derivatives[1]
F_ext_val    = F_ext(time_data)
\end{lstlisting}

\subsection*{Identification strategies}
\texttt{PyCC} supports three distinct identification strategies, categorized by the level of prior knowledge incorporated into the model: 

\subsubsection*{(i) Functional approach}
In this approach, we assume a \textit{structural skeleton} for the physics but leave the specific constitutive relations as unknown functions to be learned from data. The practitioner hypothesizes that the system consists of a velocity-dependent damping force and a position-dependent restoring force:

\begin{equation}
\ddot{x} = F_{ext}(t) - f_1(\dot{x}) - f_2(x)
\end{equation}

\begin{figure}[h]
    \centering
    \includegraphics[width=0.7\textwidth]{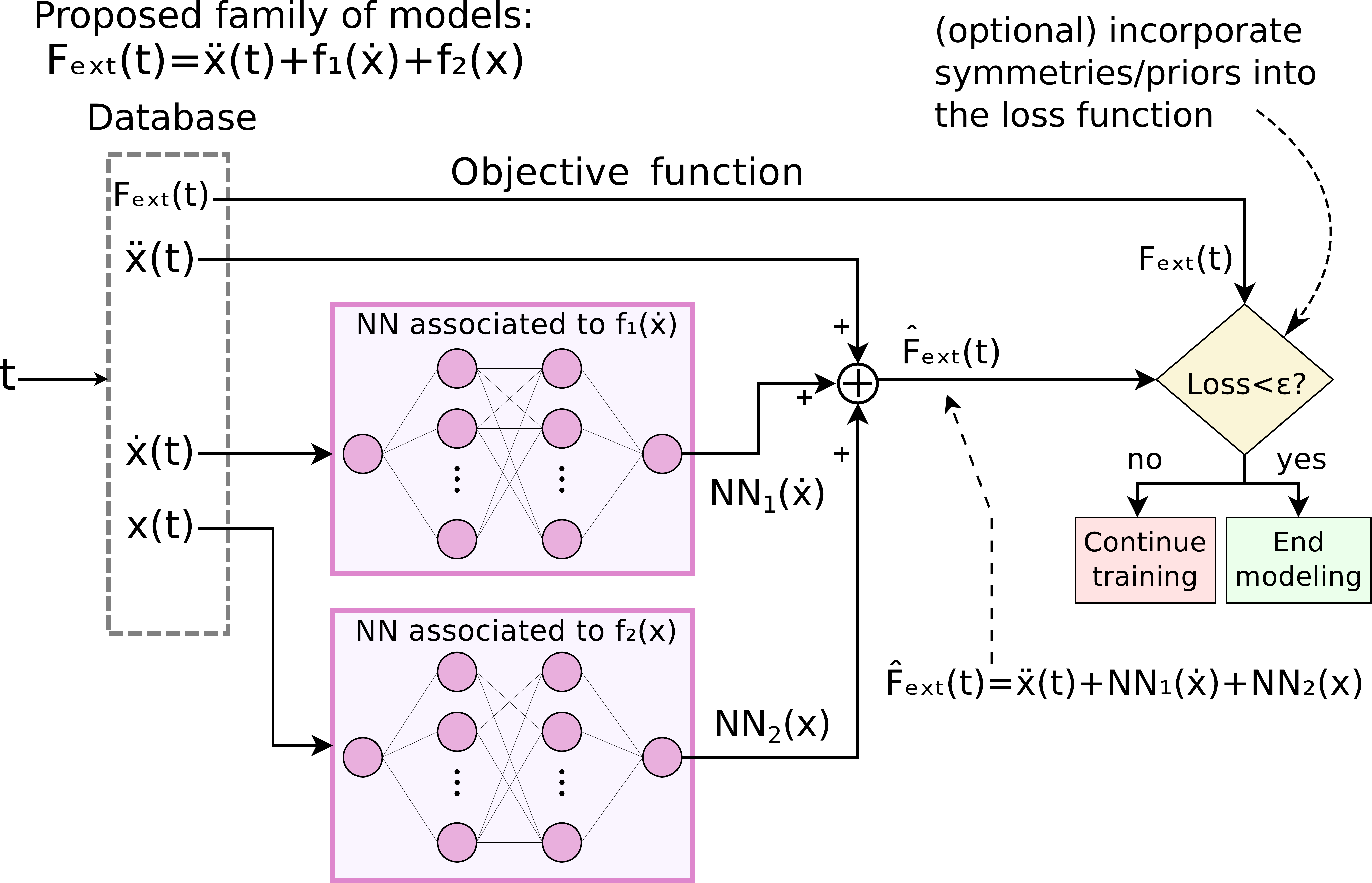}
    \caption{The architecture for a second-order system. Two independent neural networks (NN$_1$ and NN$_2$) approximate the unknown CCs. NN$_1$ sees only velocity, and NN$_2$ sees only position. Adapted with permissions from Ref.~\cite{Gonzalez2026}. }
    \label{fig:model_veloc}
\end{figure}

This family of second order systems is called as \textit{velocity-dependent friction models with external force} and has structural identifiability properties as discussed in Refs.~\cite{Gonzalez2025, Gonzalez2026}. Translating this to the state-space representation required by \texttt{PyCC}:

\begin{equation}
\begin{cases}
\dot{x}_1 = x_2 \\
\dot{x}_2 = F_{ext}(t) - f_1(x_2) - f_2(x_1)
\end{cases}
\label{eq:example_proposed_eq_functional}
\end{equation}

\textbf{The goal:} Discover the shapes of the CCs $f_1(x_2)$ and $f_2(x_1)$. These functions can be approximated using neural networks (NN), symbolic regression (SymbR), or polynomial expansions (Poly).

Figure~\ref{fig:model_veloc} presents the internal architecture that \texttt{PyCC} automatically performs when defining the skeleton of Eq.~\ref{eq:example_proposed_eq_functional} for the NN method. The following script shows the complete identification workflow using method=`NN'. 


\begin{lstlisting}
##############################################
# Defining the database for training
df = pd.DataFrame({
    'x1':x1_data,
    'x2':x2_data,
    'x1_dot':x1_dot_data,
    'x2_dot':x2_dot_data,
    'F_ext': F_ext_val
})
# Defining the proposed equations to use for identification
eqs = [
     'x1_dot = x2',
     'x2_dot = F_ext - f1(x2) - f2(x1)'
]
# In this example, we use the functional approach
# with the NN-CC method [pycc.train(method='NN')]
constraints = [ # adding prior known information
    {'constraint': 'f2(0)=0'},
    {'constraint': 'f1 odd'},
    {'constraint': 'f2 odd'},
]
# defining training parameters (optional)
params_NN = {
    'neurons': 100,
    'layers':3,
    'lr': 1e-4,
    'epochs': 2000,
    'error_threshold': 1e-6,
    'extrapolation': None,
    'device':'cpu',
    'weight_loss_param': 1e-3,
    'constraints': constraints,
}
# training the model using NN-CC method
models, evals, obtained_coefs = pycc.train(df, eqs,method='NN', params=params_NN)
# plotting obtained functions f1 and f2
x_f1_cc, f1_cc, x_f2_cc, f2_cc = evals
fig, ax = plt.subplots(1, 2, figsize=(12, 6))
ax[0].plot(x_f1_cc, f1_cc, label='$f_1$ learned NN-CC')
ax[0].plot(x_f1_cc, F1_th(x_f1_cc), '--', label='$f_1$ theory')
ax[0].set_xlabel('$x_2$')
ax[0].set_ylabel('$f_1(x_2)$')
ax[0].legend()
ax[1].plot(x_f2_cc, f2_cc, label='$f_2$ learned NN-CC')
ax[1].plot(x_f2_cc, F2_th(x_f2_cc), '--', label='$f_2$ theory')
ax[1].set_xlabel('$x_1$')
ax[1].set_ylabel('$f_2(x_1)$')
ax[1].legend()
plt.tight_layout()
plt.show()

# Print learned parameters (if any)
if obtained_coefs:
    print('\nLearned scalar parameters:')
    for name, val in obtained_coefs.items():
        print(f'{name} = {val.item():.4f}')

##############################################
# simulating forward the identified model 
# using NN-CC method [pycc.simulate(method='NN')]
params_NN_simul = {
    'models': models,
    'obtained_coefs': obtained_coefs,
    'local_funcs': {'F_ext': lambda t: F_ext(t)},
    't_span':t_span,
    'y0': y0,
    't_eval': t_eval,
    'method': 'LSODA',  
    'atol': 1e-8,
    'rtol': 1e-6,
    'check_nan': True
}
# integrating identified equations
sol,_ = pycc.simulate(eqs, method='NN', params=params_NN_simul)
time_sim=sol.t
x1_sim=sol.y[0]
x2_sim=sol.y[1]
# Plotting identified vs theoretical solution
plt.figure()
plt.plot(time_sim, x1_sim, label='x(t) simulated with NN method')
plt.plot(time_data, x1_data, label='x(t) th')
plt.xlabel('t')
plt.ylabel('x(t)')
plt.legend()
plt.show()
\end{lstlisting}

Additionally, the following script demonstrates how to use the `evals' variable to perform a post-processing workflow in which the previously obtained CCs are fitted analytically and then simulated forward using, in this example, a \texttt{pchip} interpolation.

\begin{lstlisting}
##############################################
# Post-processing the identified CCs with 
#       SymbR [pycc.post_processing()].   
print('post-SR processing for the CCs')
# Defining settings for the PySR fit
pysr_settings = {
    'niterations': 100,
    'populations': 20,
    'binary_operators': ['+', '*', '-'],
    'unary_operators': ['tanh', 'sin','cos'],
    'maxsize': 20
}
# Defining the 'params' dictionary for the post-processing function
post_process_params = {
    'evals': evals,
    'pysr': pysr_settings,
    'plot': True,  
    'n_eval': 200, 
}
# Running the post-processing
models_sr,evals_sr = pycc.post_processing(eqs, method='SymbR', params=post_process_params)

##############################################
# We can use the post-processed CCs for forward simulation
params_sim_interp = {
    'evals': evals_sr,          
    'obtained_coefs': obtained_coefs,  
    'local_funcs': {'F_ext': F_ext},
    'interp_method': 'pchip',   
    't_span': t_span,
    'y0': y0,
    't_eval': t_eval,
}
# We simulate the system by interpolating the CCs
sol, derivs = pycc.simulate(eqs, method='Interp', params=params_sim_interp)
\end{lstlisting}

\subsubsection*{(ii) Parametric Approach}
In this approach, the practitioner explicitly defines the symbolic structure of the candidate functions using parameters $\{a_i\}$, which \texttt{PyCC} then optimizes to fit the data:  

\begin{equation*}
\begin{cases}
\dot{x}_1 = x_2 \\
\dot{x}_2 = F_{ext}(t) - a_1\,x_2 - a_2\tanh(a_3\,x_2) - a_4\,x_1 - a_5\, x_1^3
\end{cases}
\end{equation*}

\textbf{The goal:} Identify the optimal scalar parameters $\{a_i\}$ that minimize the error between the model and the data. 

\begin{lstlisting}
# Defining the proposed equations to use for identification
eqs = [
     'x1_dot = x2',
     'x2_dot = F_ext - a1 x2 - a2 tanh(a3 x2) - a4 x1 - a5 x1^3'
]
\end{lstlisting}

\subsubsection*{(iii) Hybrid Approach}

\texttt{PyCC} also supports a hybrid approach that integrates functional discovery with parametric optimization. For example:

\begin{equation*}
\begin{cases}
\dot{x}_1 = x_2 \\
\dot{x}_2 = F_{ext}(t) - f_1(x_2) - a_4\, x_1 - a_5\, x_1^3
\end{cases}
\end{equation*}

\begin{lstlisting}
# Defining the proposed equations to use for identification
eqs = [
     'x1_dot = x2',
     'x2_dot = F_ext - f1(x2) - a4 x1 - a5 x1^3'
]
\end{lstlisting}

\texttt{PyCC} reserves the names $f_i$ and $a_i$ to represent functions and scalar parameters, respectively.

\subsubsection*{Using other methods}

Alternatives to \texttt{NN}, including \texttt{Poly} and \texttt{SymbR}, can be used to represent \(f_{i}\) functions with only minor code changes.
For instance, the following code shows the implementation of the \texttt{SymbR} method:


\begin{lstlisting}
params_SymbR = {
  'pysr': {
    'niterations': 100,
    'unary_operators': ['sin','cos','tanh'],
    'binary_operators': ['+','-','*'],
    'maxsize': 12,
    'populations':10,
    'model_selection': 'accuracy', 
    'verbosity': 0
  },
  'N_fit_points': 200,
  'max_iterations': 25,
}
models, evals , obtained_coefs = pycc.train(
    df=df,
    equations=equations,
    method='SymbR',
    params=params_SymbR
)

sol,_ = pycc.simulate(equations, method='SymbR', params=params_SR_simul)
\end{lstlisting}

\section*{Declaration of competing interest}
The authors declare that they have no known competing
financial interests or personal relationships that could have
appeared to influence the work reported in this paper.

\section*{Data availability}

The software package is available at: \url{https://github.com/FedejGon/pyCC.id}


\section*{AI usage disclose}
The core structure of the source code was authored manually. AI tools were subsequently employed during the development phase to assist with code refactoring and the generation of comments in the code. All AI-assisted components were rigorously reviewed, tested, and validated by the
author to ensure functional integrity. 

Additionally, AI tools were utilized during the drafting process of this manuscript to provide feedback on clarity and wording. All AI-generated suggestions were subject to critical revision and final approval by the author. 

\section*{Acknowledgments} 
The author acknowledges fruitful discussions with Luis P. Lara, Carlos E. Repetto, Bernardo J. Gómez, Rodolfo Id Betán, Ignacio Pomponio, and Luis Manuel. This work was supported by ANPCyT Project PICT-2021-I-A-01135, CONICET Project PIP 1679, and the UNR Project PID 80020190100011UR
(Argentina). 
The author acknowledges the CCT-Rosario Computational Center for the provision of computing resources and the Secretaría de Innovación, Ciencia y Tecnología (SICYT) of Argentina for access to Clementina XXI supercomputer (project PCI-91), both of which were used to develop and test this library.



\printbibliography

@book{Nayfeh_2004,
	doi = {10.1002/9783527617562},
	year = 2004,
	month = {aug},
  publisher={John Wiley \& Sons},
  address = {New York, NY},	
	author = {Ali H. Nayfeh and P. Frank Pai},
	title = {Linear and Nonlinear Structural Mechanics}
}

@article{brunton2016,
author = {Steven L. Brunton  and Joshua L. Proctor  and J. Nathan Kutz },
title = {Discovering governing equations from data by sparse identification of nonlinear dynamical systems},
journal = {Proc. Natl. Acad. Sci.},
volume = {113},
number = {15},
pages = {3932-3937},
year = {2016},
doi = {10.1073/pnas.1517384113},
}

@misc{Cranmer2023PySR,
title={Interpretable Machine Learning for Science with PySR and SymbolicRegression.jl}, 
author={Miles Cranmer},
      year={2023},
      eprint={2305.01582},
      archivePrefix={arXiv},
      primaryClass={astro-ph.IM},
}

@article{Gonzalez2023,
  title     = {Determination of the characteristic curves of a nonlinear first order system from Fourier analysis},
  author    = {Gonzalez, Federico J.},
  journal   = "Sci. Rep.",
  publisher = "Springer Science and Business Media LLC",
  volume    =  13,
  number    =  1,
  pages     = "1955",
  month     =  feb,
  year      =  2023,
  url = "",
doi =   "10.1038/s41598-023-29151-5",
  language  = "en"
}

@article{Gonzalez2024,
  title = {System identification based on characteristic curves: a mathematical connection between power series and Fourier analysis for first-order nonlinear systems},
  author = {{F. J. Gonzalez}},
  volume = {112},
  ISSN = {1573-269X},
  url = {},
  DOI = {10.1007/s11071-024-09890-4},
  number = {18},
  journal = {Nonlinear Dyn.},
  publisher = {Springer Science and Business Media LLC},
  year = {2024},
  month = jul,
  pages = {16167–16197}
}

@article{Gonzalez2025,
  title = {Interpretable neural network system identification method for two families of second-order systems based on characteristic curves},
  volume = {113},
  ISSN = {1573-269X},
  DOI = {10.1007/s11071-025-11744-6},
  number = {24},
  journal = {Nonlinear Dyn.},
  publisher = {Springer Science and Business Media LLC},
  author = {Gonzalez,  Federico J. and Lara,  Luis P.},
  year = {2025},
  month = sep,
  pages = {33063–33086}
}

@article{Gonzalez2026,
  title={Integrating prior knowledge in equation discovery: Interpretable symmetry-informed neural networks and symbolic regression via characteristic curves}, 
  author={Federico J. Gonzalez},
  year={2026},
  eprint={2601.21720},
  archivePrefix={arXiv},
  primaryClass={nlin.CD},
  url={https://arxiv.org/abs/2601.21720}, 
}

@misc{Schonlieb2025,
      title={Data-driven approaches to inverse problems}, 
      author={Carola-Bibiane Schönlieb and Zakhar Shumaylov},
      year={2025},
      eprint={2506.11732},
      archivePrefix={arXiv},
      primaryClass={math.NA},
      url={https://arxiv.org/abs/2506.11732}, 
}

@article{Dietrich2025,
  title = {Scientific machine learning},
  volume = {72},
  ISSN = {1432-1815},
  DOI = {10.1007/s00591-025-00399-4},
  number = {2},
  journal = {Mathematische Semesterberichte},
  publisher = {Springer Science and Business Media LLC},
  author = {Dietrich,  Felix and Schilders,  Wil},
  year = {2025},
  month = Sept,
  pages = {89–115}
}

@incollection{Paszke2019,
title = {PyTorch: An Imperative Style, High-Performance Deep Learning Library},
author = {Paszke, Adam and Gross, Sam and Massa, Francisco and Lerer, Adam and Bradbury, James and Chanan, Gregory and Killeen, Trevor and Lin, Zeming and Gimelshein, Natalia and Antiga, Luca and Desmaison, Alban and Kopf, Andreas and Yang, Edward and DeVito, Zachary and Raison, Martin and Tejani, Alykhan and Chilamkurthy, Sasank and Steiner, Benoit and Fang, Lu and Bai, Junjie and Chintala, Soumith},
booktitle = {NeurIPS 32},
editor = {H. Wallach and H. Larochelle and A. Beygelzimer and F. d\textquotesingle Alch\'{e}-Buc and E. Fox and R. Garnett},
pages = {8024--8035},
year = {2019},
publisher = {Curran Associates, Inc.},
}

@misc{Chen2018,
      title={Neural Ordinary Differential Equations}, 
      author={Ricky T. Q. Chen and Yulia Rubanova and Jesse Bettencourt and David Duvenaud},
      year={2018},
      eprint={1806.07366},
      archivePrefix={arXiv},
      primaryClass={cs.LG},
      url={https://arxiv.org/abs/1806.07366}, 
}

@article{Raissi2019,
title = {Physics-informed neural networks: A deep learning framework for solving forward and inverse problems involving nonlinear partial differential equations},
journal = {Journal of Computational Physics},
volume = {378},
pages = {686-707},
year = {2019},
issn = {0021-9991},
doi = {10.1016/j.jcp.2018.10.045},
author = {M. Raissi and P. Perdikaris and G.E. Karniadakis},
}
\end{document}